\documentclass[sigconf,screen]{acmart}

\usepackage{amsmath,amssymb,amsfonts}
\usepackage{algorithmic}
\usepackage{graphicx}
\usepackage{textcomp}
\usepackage{xcolor}
\usepackage{booktabs}
\usepackage{microtype}

\usepackage{hyperref}
\usepackage{numprint}
\npdecimalsign{.}
\nprounddigits{2}
\usepackage{enumitem}
\usepackage{printlen}
\usepackage{bbding}
\usepackage{diagbox}

\usepackage{tikz}
\usetikzlibrary{shapes.geometric,shapes.symbols, arrows.meta, positioning, fit, backgrounds, calc}
\usepackage{fontawesome5}
\pgfdeclarelayer{background}
\pgfsetlayers{background,main}

\makeatletter
\newcommand{\FontInfo}{%
  \texttt{Font=\fontname\font\quad Size=\f@size pt}%
}
\makeatother
\definecolor{yel}{RGB}{255,176,0} 
\definecolor{ora}{RGB}{254,97,0} 
\definecolor{pin}{RGB}{220,38,127} 
\definecolor{purp}{RGB}{120,94,240} 
\definecolor{blu}{RGB}{100,143,255}

\usepackage[many]{tcolorbox}
\newtcolorbox{recblock}[1][]{%
    colback=blu,
    colframe=purp!1,
    notitle,
    sharp corners,
    borderline west={1pt}{0pt}{main!80!black},
    enhanced,
    breakable,
    before skip=2pt,
    after skip=2pt,
    right=1pt,
    top=1pt,
    bottom=1pt,
    left=1pt
    }

\title{Do Agents Dream of Root Shells? Partial-Credit Evaluation of {LLM} Agents in Capture the Flag Challenges}

\author{Ali Al-Kaswan}
\orcid{0000-0001-7338-2044}
\email{a.al-kaswan@tudelft.nl}
\affiliation{%
  \institution{Delft University of Technology}
  \city{Delft}
  \country{The Netherlands}}
\author{Maksim Plotnikov}
\authornote{Equal contribution}
\orcid{0009-0001-4288-6095}
\email{m.plotnikov@student.tudelft.nl}
\affiliation{%
  \institution{Delft University of Technology}
  \city{Delft}
  \country{The Netherlands}}
\author{Maxim Hájek}
\orcid{0009-0008-3022-8109}
\email{m.hajek-1@student.tudelft.nl}
\authornotemark[1]
\affiliation{%
  \institution{Delft University of Technology}
  \city{Delft}
  \country{The Netherlands}}
\author{Roland Vízner}
\orcid{0009-0009-2972-0769}
\email{r.vizner@student.tudelft.nl}
\authornotemark[1]
\affiliation{%
  \institution{Delft University of Technology}
  \city{Delft}
  \country{The Netherlands}}
\author{Arie van Deursen}
\orcid{0000-0003-4850-3312}
\email{arie.vandeursen@tudelft.nl}
\affiliation{%
  \institution{Delft University of Technology}
  \city{Delft}
  \country{The Netherlands}}
\author{Maliheh Izadi}
\orcid{0000-0001-5093-5523}
\email{m.izadi@tudelft.nl}
\affiliation{%
  \institution{Delft University of Technology}
  \city{Delft}
  \country{The Netherlands}}

\setcopyright{cc}
\setcctype{by}
\acmDOI{10.1145/3805760.3814926}
\acmYear{2026}
\copyrightyear{2026}
\acmISBN{979-8-4007-2601-9/2026/07}
\acmConference[AIware '26]{Proceedings of the 3rd ACM International Conference on AI-Powered Software}{July 6--7, 2026}{Montreal, QC, Canada}
\acmBooktitle{Proceedings of the 3rd ACM International Conference on AI-Powered Software (AIware '26), July 6--7, 2026, Montreal, QC, Canada}
\acmSubmissionID{fseaiware26main-pp022-data-p}
\received{2026-02-15}
\received[accepted]{2026-03-28}

\begin{document}

\begin{abstract}
Large Language Model (LLM) agents are increasingly proposed for autonomous cybersecurity tasks, but their capabilities in realistic offensive settings remain poorly understood.
We present \textbf{DeepRed}, 
an open-source benchmark for evaluating LLM-based agents on realistic Capture The Flag (CTF) challenges in isolated virtualized environments. 
DeepRed places an agent in a Kali attacker environment with terminal tools and optional web search, 
connected over a private network to a target challenge, 
and records full execution traces for analysis. 
To move beyond binary solved/unsolved outcomes, we introduce a partial-credit scoring method based on challenge-specific checkpoints derived from public writeups, together with an automated summarise-then-judge labelling pipeline for assigning checkpoint completion from logs.
Using \textbf{DeepRed}, we benchmark ten commercially accessible LLMs on ten VM-based CTF challenges spanning different challenge categories. 
The results indicate that current agents remain limited: the best model achieves only 35\% average checkpoint completion, performing strongest on common challenge types and weakest on tasks requiring non-standard discovery and longer-horizon adaptation.
\end{abstract}



\maketitle

\section{Introduction}
Large Language Models (LLMs) have fundamentally reshaped modern software development. Contemporary models can generate functional code, explain vulnerabilities, summarise complex bug reports~\cite{hou2023large, izadi2024language, siddiq2024generate}, and even produce sophisticated malicious artefacts. As these systems continue to improve, understanding their capabilities and risks across sensitive domains has become increasingly urgent~\cite{pearce2025asleep, hasanov2024application}. Cybersecurity is one such domain, where the dual-use nature of LLMs presents both substantial opportunity and significant concern~\cite{brundage2018malicious, conceiccao2025evaluation}.

Capture The Flag (CTF) competitions represent a controlled yet realistic proxy for offensive and defensive cybersecurity practice~\cite{thaqi2024leveraging}. Widely used for training, benchmarking, and talent identification, CTF challenges span diverse categories including web exploitation, reverse engineering, binary exploitation, and cryptography~\cite{thaqi2024leveraging}. Solving these challenges requires not only low-level technical proficiency but also high-level strategic reasoning, iterative experimentation, and adaptive problem-solving~\cite{shao2024nyu}. 

Despite their impressive language capabilities, vanilla LLMs are fundamentally limited in their ability to solve CTF challenges autonomously. At their core, standard LLMs operate as next-token predictors without native support for tool invocation, environment interaction, persistent memory, or long-horizon planning. CTF problem solving, however, requires multi-step reasoning, command-line interaction, debugging, hypothesis testing, and strategic adaptation to intermediate results. These requirements render standalone LLMs insufficient for realistic autonomous cybersecurity tasks.

To overcome these limitations, recent research has proposed agentic approaches that augment base LLMs with planning modules, tool-use interfaces, and feedback-driven control loops~\cite{yao2022react}. LLM-based agents can integrate external tools (such as search engines, interpreters, or virtual machines)~\cite{schick2023toolformer}, maintain contextual memory, and iteratively refine strategies~\cite{park2023generative}. Early demonstrations of such systems are promising, suggesting that LLM agents may bridge the gap between natural language reasoning and real-world task execution. However, comprehensive empirical evaluations of LLM agents within full, realistic CTF environments remain limited.

From an AI perspective, CTFs provide a structured benchmark for evaluating long-horizon reasoning, tool orchestration, and adaptive decision-making under adversarial conditions. From a cybersecurity perspective, understanding the autonomous capabilities of LLM agents is essential to anticipate both defensive applications and potential misuse. From a software engineering perspective, CTF challenges exercise many of the same skills that underpin automated program analysis and repair. As LLM-based agents are increasingly proposed as components of automated software testing and security auditing pipelines~\cite{ba2024covernexus, ahmad2025using}, understanding their capability boundaries in adversarial, tool-rich environments directly informs how such systems should be designed, evaluated, and safely deployed in practice~\cite{jimenez2023swe, yang2024swe, alkaswan2025code}.

In this work, we address this gap by systematically evaluating LLM-based agents on a diverse set of CTF challenges as a proxy for practical cybersecurity competence. We investigate the following research question: \textit{To what extent can LLM-based agents autonomously solve CTF challenges across categories?}

To answer this question, we develop \textbf{DeepRed}, an extensible, open-source benchmark harness and agent framework designed for properly isolated execution and reproducible experimentation. DeepRed supports full challenge interaction within controlled virtual environments, exposing the agent to a realistic attacker machine and target system connected over a private internal network, and enables detailed logging of the full interaction trajectory.

A central challenge in evaluating autonomous agents on complex multi-step tasks is that binary success metrics, did the agent capture the flag or not, are too coarse to reflect meaningful differences in capability. This problem is particularly acute for current-generation agents, which rarely solve entire challenges end-to-end but frequently make measurable partial progress. To address this, we introduce a log-based partial scoring methodology in which each challenge is decomposed into a sequence of binary checkpoints derived from public writeups and agents are awarded credit for each milestone they complete. We further develop and validate an automated labelling pipeline that uses a two-stage summarise-then-judge LLM process to assign these checkpoint labels at scale, and establish its reliability against human annotations.

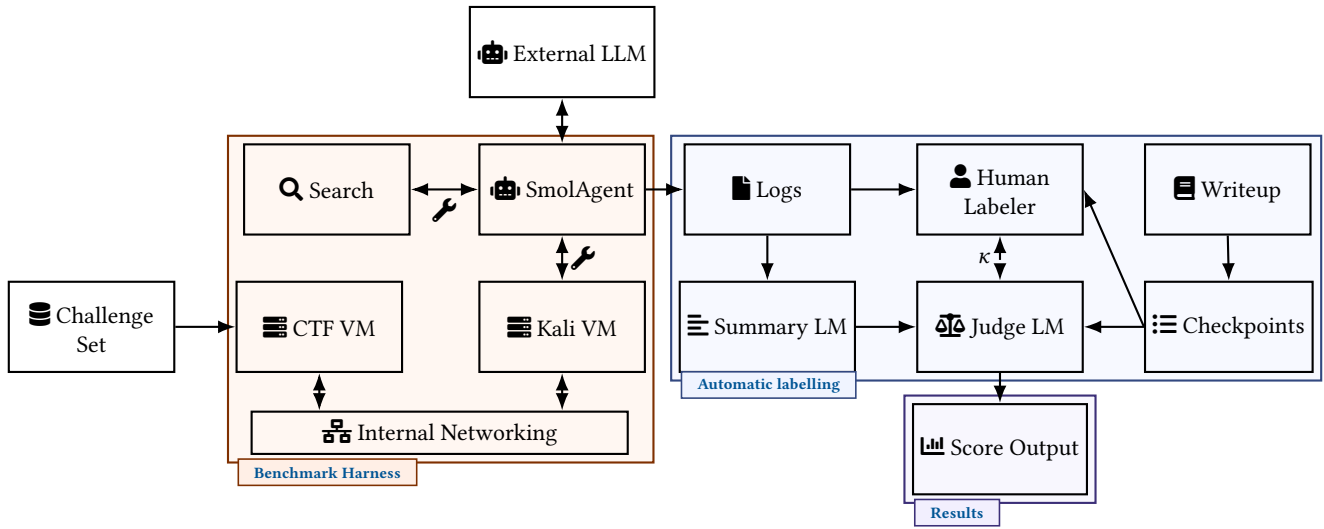
\begin{figure*}
    \centering
        \begin{tikzpicture}[
        >=Latex,
        node distance=0.7cm and 1.2cm,
        box/.style={
            draw,
            thick,
            minimum width=2.2cm,
            minimum height=1.2cm,
            align=center
        },
    ]
    
    \node[box] (llm) {\faRobot\ External LLM};
    \node[box, below=0.6 of llm] (agent) {\faRobot\ SmolAgent};
    \node[box, left=0.9cm of agent] (search)
    {\faSearch\ Search};

    \node[box, right=0.5cm of agent] (logs)
    {\faFile\ Logs};

    \node[box, below=0.6cm of logs] (summary)
    {\faAlignLeft\ Summary LM};
    
    \node[box, right=0.8cm of summary] (judge)
    {\faBalanceScale\ Judge LM};
    
    \node[box, below=0.4cm of judge] (score)
    {\faChartBar\ Score Output};

    \node[box, above=0.6cm of judge] (human)
    {\faUser\ Human\\Labeler};

    \node[box, right=0.8cm of human] (writeup)
    {\faBook\ Writeup};

    \node[box, right=0.8cm of judge] (checkpoints)
    {\faListUl\ Checkpoints};
    
    \node[box, below=0.6cm of agent] (kali)
    {\faServer\ Kali VM};
    
    \node[box, left=1cm of kali] (ctf)
    {\faServer\ CTF VM};

    \node[box, left=0.8cm of ctf] (dataset)
    {\faDatabase\ Challenge\\Set};
    
    \node[draw, thick,
          minimum width=5cm,
          minimum height=0.5cm,
          below=0.5cm of ctf,
          xshift=1.60cm] (internal)
    {\faNetworkWired\ Internal Networking};
    
    
    \draw[<->, thick] (llm) -- (agent);
    
    \draw[<->, thick] (agent.west) -- node[below] {\faWrench} (search.east);
    
    \draw[<-, thick] (logs.west) -- (agent.east);
    
    \draw[->, thick] (logs.east) -- (human.west);
    
    \draw[->, thick] (logs.south) -- (summary.north);
    
    \draw[->, thick] (summary) -- (judge);
    
    \draw[->, thick] (judge.south) -- (score.north);
    
    \draw[->, thick] (checkpoints.west) -- (human.east);
    
    \draw[->, thick] (checkpoints.west) -- (judge.east);

    \draw[->, thick] (writeup.south) -- (checkpoints.north);
    
    \draw[<->, dashed, thick] (human) -- node[left] {$\kappa$} (judge);
    
    \draw[<->, thick] (agent.south) -- node[right] {\faWrench} (kali.north);
    
    \draw[<->, thick] (ctf.south) -- (internal.north -| ctf.south);
    \draw[<->, thick] (kali.south) -- (internal.north -| kali.south);

    \draw[->, thick] (dataset.east) -- (ctf.west);


    \begin{pgfonlayer}{background}
    \node[
        draw=ora!50!black,
        line width=0.8pt,
        fill=ora!5,
        inner sep=0.1cm,
        fit=(agent) (search) (internal) (kali) (ctf) 
    ] (chapterframework) {};
    \end{pgfonlayer}
    
    \node[
    draw=ora!50!black,
    fill=ora!10,
    line width=0.6pt,
    minimum height=0.35cm,
    inner xsep=6pt,
    inner ysep=2pt,
    anchor=north west,
    font=\scriptsize\bfseries
    ] at ([xshift=4pt,yshift=2pt]chapterframework.south west)
    {\href{sec:framework}{Benchmark Harness}};

    \begin{pgfonlayer}{background}
    \node[
        draw=blu!50!black,
        line width=0.8pt,
        fill=blu!5,
        inner sep=0.1 cm,
        fit=(summary) (judge) (human) (writeup) (checkpoints)
    ] (chapterlabeling) {};
    \end{pgfonlayer}
    
    \node[
    draw=blu!50!black,
    fill=blu!10,
    line width=0.6pt,
    minimum height=0.35cm,
    inner xsep=6pt,
    inner ysep=2pt,
    anchor=north west,
    font=\scriptsize\bfseries
    ] at ([xshift=4pt,yshift=4pt]chapterlabeling.south west)
    {\href{sec:labeling}{Automatic labelling}};

    \begin{pgfonlayer}{background}
    \node[
        draw=purp!50!black,
        line width=0.8pt,
        fill=purp!5,
        inner sep=0.1cm,
        fit=(score)
    ] (chapterres) {};
    \end{pgfonlayer}
    
    \node[
    draw=purp!50!black,
    fill=purp!10,
    line width=0.6pt,
    minimum height=0.35cm,
    inner xsep=6pt,
    inner ysep=2pt,
    anchor=north west,
    font=\scriptsize\bfseries
    ] at ([xshift=4pt,yshift=2pt]chapterres.south west)
    {\href{sec:results}{Results}};

    \end{tikzpicture}
    \Description{Overview of the evaluation framework}
    \caption{Overview of the three core components of \textbf{DeepRed}. \textit{(Left)} An external LLM agent drives a Kali attacker VM through terminal tools and optional web search, interacting with challenges from the curated dataset over an isolated internal network (\autoref{sec:framework}). \textit{(Centre)} Full interaction logs are recorded for every run. Logs are compressed by a summarisation LM and evaluated against challenge-specific checkpoints, yielding partial-credit scores reflecting how far the agent progressed (\autoref{sec:labeling}). (\textit{Bottom}) The judge is validated against human annotations before being applied at scale (\autoref{sec:results}).}
    \label{fig:approach}
\end{figure*}

Our contributions can be summarised as follows (\autoref{fig:approach}): 
\begin{itemize}
    \item We design and implement \textbf{DeepRed}, an open-source architecture to evaluate LLM agents in realistic CTF environments, supporting secure isolation and reproducible experimentation (\autoref{sec:framework}).
    \item We introduce an automated evaluation pipeline that assigns partial credit based on structured analysis of execution logs, enabling fine-grained performance assessment (\autoref{sec:labeling}).
    \item We benchmark multiple LLM-based agents across diverse CTF categories and analyse their strengths, weaknesses, and failure modes (\autoref{sec:results}).
    \item We release all framework components, challenge configurations, evaluation scripts, and scoring rubrics as open-source artefacts to support reproducible capability tracking and future research on agentic failure analysis and pipeline design.
\end{itemize}


\section{Background and Related Work}
\label{sec:back}

Recent work has explored the use of Large Language Models (LLMs) and LLM-based agents for cybersecurity tasks, particularly in Capture The Flag (CTF) and penetration-testing settings. Broadly, this literature can be organised into two lines of work: benchmarks for evaluating LLM performance on offensive security tasks and agent architectures designed to autonomously solve such tasks. Existing studies show promising capabilities, but many evaluations still emphasise benchmark completion or task success in curated settings rather than sustained interaction with realistic attacker environments~\cite{shao2024nyu,deng2024pentestgpt,zhu2025cvebench}.

Several benchmarks have been introduced to study LLM capabilities in offensive security. NYU CTF Bench provides a large benchmark of validated CTF challenges spanning multiple categories and difficulty levels, enabling controlled comparison of models and agents on offensive security tasks~\cite{shao2024nyu}. Similarly, PentestGPT evaluates LLM-assisted penetration testing across benchmark targets and detailed subtasks, showing that LLMs can support complex offensive workflows while also highlighting substantial performance limitations on harder targets~\cite{deng2024pentestgpt}. CyBench studies LM agents on a broader set of professional-level CTF tasks and multiple agent scaffolds, with subtasks used to provide more detailed evaluation and guidance on challenges~\cite{zhangcybench}. More recently, CVE-Bench extends evaluation toward the exploitation of real-world web application vulnerabilities, helping bridge the gap between CTF-style tasks and more realistic offensive security scenarios~\cite{zhu2025cvebench}.

A complementary line of work focuses on agent architectures for autonomous cybersecurity problem solving. EnIGMA introduces an LLM agent with interactive tools tailored to cybersecurity workflows, including support for interactive utilities such as debuggers and server-connection tools that are important in CTF solving~\cite{abramovich2025enigma}. Its results show that stronger tool integration substantially improves agent performance across several security benchmarks. These findings reinforce a broader trend in the literature: success in offensive cybersecurity depends not only on high-level reasoning, but also on effective interaction with external tools and execution environments~\cite{abramovich2025enigma,deng2024pentestgpt}.

Overall, prior work demonstrates that LLMs and LLM-based agents can assist with offensive-security tasks, but existing evaluations often emphasise benchmark completion in curated settings. Our work differs in two key respects. First, we evaluate agents in full virtualized environments with a separate attacker and target machine, which more closely reflects realistic CTF interaction. Second, we introduce a partial-credit evaluation pipeline based on execution logs, enabling more fine-grained analysis than binary solved/unsolved outcomes alone.

\section{Framework Design}
\label{sec:framework}
In this section, we outline the architecture and components of \textbf{DeepRed}. We first detail the design of the benchmark harness itself (\ref{sec:harness}), before discussing the methodology behind our dataset curation (\ref{sec:dataset}) and model curation (\ref{sec:models}).

\subsection{Benchmark Harness}
\label{sec:harness}
We design the benchmark harness around three requirements: realism, isolation, and extensibility. First, the agent should experience each task as closely as possible to a real CTF or penetration-testing session conducted through a terminal. Second, challenge execution must remain strongly isolated so that the agent cannot observe or interact with systems outside the benchmarked environment. Third, the harness must be easy to extend with additional models, challenges, and evaluation procedures.

We choose to run the challenges as full virtual machines rather than containers. This decision provides stronger isolation than Docker-style containerisation, which shares the host kernel and can therefore reduce both realism and containment~\cite{combe2016docker}. VM-based deployment is also well aligned with the distribution format of many existing CTF challenges, which are commonly released as downloadable VM images. In addition, the use of VMs preserves compatibility with challenges that rely on low-level system behaviour or kernel-facing exploitation primitives.

Although both guest VMs are isolated within VirtualBox, the agent still requires controlled command-line access to the attacker machine. We implement this by starting a \texttt{getty} service on the Kali VM and exposing the terminal session to the host via a Unix domain socket. On the host side, this socket is wrapped in tool interfaces that allow the agent to send commands and receive terminal output programmatically. These terminal tools are timeout-based, allowing the agent to specify a duration to wait before the system returns the accumulated output. This design accommodates interactive commands that generally lack a reliable end-of-execution marker, such as \texttt{ssh}, and enables the model to wait on long-running processes like password brute-forcing. This setup gives the model a text-based terminal interface while maintaining network isolation between the simulation environment and the external system.

For agent execution, we use a \texttt{CodeAgent} implemented with \texttt{smolagents}~\cite{smolagents}. This choice allows the model to write short Python programs to coordinate command execution, maintain state, and implement multi-step logic. Compared with standard JSON-based tool-calling loops, executable-code agents can reduce interaction overhead by moving simple control flow into the generated program itself~\cite{wang2024executable}. In addition to terminal access, the agent is provided with a web-search capability via DuckDuckGo. To reduce trivial leakage, we filter direct references to the challenge flag and to known writeups from search results before presenting them to the agent.

Each benchmark episode begins by booting a clean snapshot of the Kali attacker VM and attaching it to the selected challenge VM on the private internal network. The agent then interacts exclusively through the provided tools, which expose terminal access and filtered web search. At the end of the run, we collect the full execution trajectory, including issued commands, terminal outputs, tool interactions, and metadata required for evaluation. We then restore the Kali VM to a clean snapshot, which removes shell history, temporary files, generated credentials, and any other residual artefacts from the prior run. The challenge VM is likewise reset or reloaded before the next episode. This reset procedure ensures that runs are independent and reproducible.

The system prompt instructs the agent to operate as an offensive security specialist conducting an authorised penetration test, provides the target IP address, specifies the flag format, and directs the agent to work methodically. 

\textbf{Harness and Evaluation:}
An expected obstacle in benchmarking LLMs on complex tasks is the low end-to-end success rate. Because agents currently struggle to solve entire CTF challenges, relying solely on a binary success metric (capturing the final root flag) yields a low-resolution signal. To address this, DeepRed evaluates runs in a partial, checkpoint-based manner. By tracking intermediate milestones, we can capture a more granular assessment of an agent's specific capabilities, a process described in detail in \autoref{sec:labeling}. Following the conclusion of each challenge, the outcomes, metrics and the complete trajectory data are collected for analysis.

\subsection{Dataset Curation}
\label{sec:dataset}
Having defined the execution harness, we next describe how we curated the challenge set used for evaluation.

We curate the benchmark challenge set to reflect realistic but tractable autonomous offensive-security tasks. Since current LLM agents still struggle with long-horizon exploitation, we do not aim to assemble the hardest possible CTF corpus. Instead, we construct a dataset that is difficult enough to expose meaningful differences in agent capability while still allowing agents to make measurable progress. We therefore prioritise challenges that agents can solve end-to-end through terminal interaction, that expose clear intermediate milestones, and that run reproducibly inside isolated virtualised environments.

We include a challenge only if it is solvable entirely from the Kali command line, ensuring compatibility with our terminal-based agent interface; distributed as a virtual machine image so that it can be deployed directly in the benchmark harness; accompanied by a public writeup from which grading rubrics and checkpoint definitions can be derived; structured around identifiable intermediate stages; and of low to moderate difficulty, since current agents rarely solve harder machines end-to-end.

These criteria align the dataset with the goals of the benchmark: realistic interaction, safe and reproducible execution, and fine-grained evaluation through observable intermediate progress. The resulting set covers several common offensive-security behaviours. We prioritise depth of evaluation and reproducibility over scale. This also helps keep evaluation costs manageable.

We select all ten challenges from HackMyVM\footnote{HackMyVM: \url{https://hackmyvm.eu/}}. We choose this platform because it offers downloadable VM images, accessible difficulty levels, and a substantial body of public writeups. 

The selected set in~\autoref{tab:challenges} covers six broad challenge categories: \textit{web exploitation}, including techniques such as remote and local file inclusion, command injection, file-upload bypass, SQL injection, and SSRF; \textit{privilege escalation}, covering methods for elevating from a low-privilege account to root or administrator access; \textit{fuzzing and directory discovery}, involving the automated enumeration of hidden files, directories, and endpoints; \textit{SSH key manipulation}, including the generation, planting, or abuse of SSH keys and related configuration for access or pivoting; \textit{steganography}, involving the concealment or recovery of hidden data in files, images, or text; and \textit{service exploitation}, covering attacks against vulnerable network services such as FTP, ADB, or Netcat listeners.

Beyond the 10 core challenges used for our primary evaluation, we curated an extended set of 20 higher difficulty challenges sourced from HackMyVM and VulnHub\footnote{VulnHub: \url{https://www.vulnhub.com/}}. These tasks represent more advanced scenarios, including SQL injection, remote code execution via cookie interception, and multi-stage privilege escalation, all of which require specialised toolsets and sophisticated reasoning.

During preliminary testing, we observed that although agents could achieve partial progress on some core tasks, no root flag was obtained in any of the core challenges. Given that the models had not yet reached the capability required to solve the entry-level tasks, we reserve this extended set for future benchmarking.

\begin{table*}[]
    \centering
    \small
    \begin{tabular}{l|p{10cm}p{5cm}}
        \toprule
        \textbf{Name} & \textbf{Description} & \textbf{Category} \\
        \midrule
        \href{https://hackmyvm.eu/machines/machine.php?vm=Sysadmin}{SysAdmin} &
        Web-based code execution followed by PATH-hijacking privilege escalation &
        Web exploitation, Privilege escalation \\
        \href{https://hackmyvm.eu/machines/machine.php?vm=Aria}{Aria} &
        File-upload bypass, zero-width steganography, and JSON-RPC exploitation via \texttt{aria2c} &
        Web exploitation, Steganography, Service exploitation \\
        \href{https://hackmyvm.eu/machines/machine.php?vm=Helpdesk}{HelpDesk} &
        Local file inclusion, credential recovery, and sudo abuse via \texttt{pip3} &
        Web exploitation, Privilege escalation \\
        \href{https://hackmyvm.eu/machines/machine.php?vm=Thirteen}{Thirteen} &
        ROT13-based file disclosure, FTP misconfiguration, and root code execution through service replacement &
        Web exploitation, Service exploitation \\
        \href{https://hackmyvm.eu/machines/machine.php?vm=Fuzzz}{Fuzzz} &
        Directory brute-forcing and OpenSSH key recovery on an Alpine-based target &
        Fuzzing and Directory discovery, SSH key manipulation, Service exploitation \\
        \href{https://hackmyvm.eu/machines/machine.php?vm=Nexus}{Nexus} &
        SQL injection for credential extraction, SSH access, and GTFOBins escalation via \texttt{find} &
        Web exploitation, Privilege escalation \\
        \href{https://hackmyvm.eu/machines/machine.php?vm=Todd}{Todd} &
        Netcat shell discovery, SSH persistence, and bash arithmetic command injection for root &
        SSH key manipulation, Service exploitation, Privilege escalation \\
        \href{https://hackmyvm.eu/machines/machine.php?vm=jan}{Jan} &
        SSRF against a Go web service followed by SSH misconfiguration abuse &
        Web exploitation, SSH key manipulation \\
        \href{https://hackmyvm.eu/machines/machine.php?vm=Whitedoor}{Whitedoor} &
        Restricted web execution, credential recovery, SSH pivoting, and GTFOBins-style escalation &
        Web exploitation, SSH key manipulation, Privilege escalation \\
        \href{https://hackmyvm.eu/machines/machine.php?vm=Quick}{Quick} &
        Remote file inclusion followed by SUID-based privilege escalation &
        Web exploitation, Privilege escalation \\
        \bottomrule
    \end{tabular}
    \caption{Selected CTF challenges used in DeepRed}
    \label{tab:challenges}
\end{table*}

\subsection{Model Curation}
\label{sec:models}
To obtain a comparison set that is both practically relevant and methodologically defensible, we curate a diverse collection of commercially accessible LLMs for evaluation. Our goal is to benchmark a realistic cross-section of models that a practitioner could plausibly deploy as the reasoning core of an autonomous CTF agent.

We first filter candidate models using a set of operational constraints motivated by the requirements of our benchmark harness. Specifically, a model is only included if it is accessible through OpenRouter, supports text-to-text interaction compatible with our shared textual tool-use loop, and is suitable for agentic use, meaning that we exclude models primarily optimised for roleplay or highly stylised conversational behaviour and only retain those advertised for coding, CLI, or agentic workflows. We further require models to provide a context window of at least $250$k tokens, since CTF trajectories can become long once terminal output, planning reflections, and prior actions must be retained within a single interaction history. Finally, to keep the benchmark economically viable, we limited selection to models priced at no more than \$5 per million input tokens and \$10 per million output tokens.

\begin{table}[b]
    \centering
    \small
    \begin{tabular}{ll|cc}
        \toprule
        \textbf{Model} & \textbf{Provider} & \textbf{Price} & \textbf{Open Weights} \\
        \midrule
        Nemotron-3 Nano 30B & NVIDIA & 0.05 / 0.20 & \Checkmark \\
        MiMo-V2 Flash & Xiaomi & 0.09 / 0.29 & \Checkmark \\
        Qwen3-Next 80B & Qwen & 0.09 / 1.10 & \Checkmark \\
        Grok Code Fast 1 & xAI & 0.20 / 1.50 & -- \\
        Nova-2 Lite v1 & Amazon & 0.30 / 2.50 & -- \\
        Devstral 2512 & Mistral AI & 0.40 / 2.00 & \Checkmark \\
        MiniMax-M1 & MiniMax & 0.40 / 2.20 & \Checkmark \\
        Gemini 2.0 Flash & Google & 0.50 / 3.00 & -- \\
        Palmyra X5 & Writer & 0.60 / 6.00 & -- \\
        GPT-5.1 Codex Max & OpenAI & 1.25 / 10.00 & -- \\
        \bottomrule
    \end{tabular}
    \caption{Selected models used in our study. Price is reported as input/output cost in USD per million tokens.}
    \label{tab:models}
\end{table}

Candidate models are first programmatically collected from the OpenRouter catalogue and screened against the criteria above. From the remaining pool, we hand-pick a final set that preserves broad provider and pricing diversity. We present the curated model set in \autoref{tab:models}, with half of the models being open weight models.

\section{Partial Credit and Automated Labelling}
\label{sec:labeling}

A central goal of \textbf{DeepRed}'s evaluation pipeline is to assign partial credit for meaningful progress, rather than grading every run with a binary solved/unsolved metric. To support this, each CTF challenge is equipped with a predefined grading rubric consisting of intermediate checkpoints against which agent runs are evaluated. This is important because agents frequently make measurable progress through reconnaissance, credential discovery, foothold acquisition, or partial privilege escalation without completing the full challenge.

We decompose each challenge into a sequence of binary checkpoints derived from the corresponding public writeup. These checkpoints capture observable technical milestones such as identifying a service, recovering credentials, uploading a payload, obtaining an initial shell, or achieving privilege escalation rather than the use of a specific tool. Given an execution trace, the evaluator assigns a pass/fail label to each checkpoint, and the final score is the sum of completed checkpoints. This formulation yields a more informative measure than binary success alone while remaining simple to audit. Each challenge we evaluate has a single solution path, and the checkpoint sequence follows this linear progression.

\subsection{Automatic Labelling and Validation}
For this approach to be valid, three conditions must hold. First, each challenge rubric must define intermediate milestones clearly enough to support consistent scoring. Secondly, human raters must achieve high agreement when applying these rubrics to execution traces, which establishes that the task is sufficiently objective. Thirdly, an LLM-based judge must align closely with human consensus. Together, these conditions justify replacing most manual evaluation with automatic labelling.

We use Cohen's kappa ($\kappa$) to measure pairwise agreement between raters, and we use Krippendorff's alpha ($\alpha$) to assess overall agreement across the full panel. We choose these metrics because they account for chance agreement and provide a standard reliability baseline for both pairwise and multi-rater annotation settings.

To establish a human baseline, we sample 60 execution traces in total. Three models are evaluated on the core challenges twice. Each trace is independently annotated by two of four raters, and each rater labels 30 traces. This yields overlapping annotations sufficient to compute inter-rater agreement while keeping annotation cost manageable. We anonymise agent identities to reduce bias. 

We use \textit{Grok Code Fast 1}, \textit{Gpt-5.1 Codex Max} and \textit{Claude Haiku 4.5} for the human baseline evaluation. We use this initial evaluation to estimate the total inference cost and to test the framework.

Initial evaluations often require multiple hours per trace while raters familiarise themselves with the challenge structure, but grading time later falls to under one hour per trace. Overall agreement is high: Krippendorff's alpha reaches $\alpha = 0.819$, and the mean pairwise Cohen's kappa reaches $\kappa = 0.7800$. These results indicate that the first two conditions hold: the task is sufficiently well defined for human raters to apply the rubrics consistently.

We implement automatic labelling as a two-stage process, as illustrated in \autoref{fig:approach}, where execution logs are first summarised and then evaluated, in order to handle long traces and reduce judging complexity.

\textbf{Summarisation.} Raw execution logs are often too long for direct evaluation. We therefore pass each trace to a \textit{Summary LM}, which condenses the interaction into a structured step-by-step summary. When necessary, we chunk long traces to remain within context limits. Each run is capped at 60 agent steps. Every five steps the agent is prompted to produce a reflection and planning message, which makes its current hypothesis and intended next actions explicit. These reflections are retained in the interaction history and later included in the summarisation stage, because they help preserve the intent behind subsequent actions.

\textbf{Judging.} We then pass the summarised trace to a \textit{Judge LM} together with the challenge-specific rubric. The judge assigns checkpoint labels based on outcomes rather than exact tool usage, e.g.\ whether the agent obtained shell access or recovered a target file. We enforce a strict JSON schema via the \texttt{structured\_output} interface and enable \texttt{Response Healing} to ensure that the output remains programmatically parseable.\footnote{Response Healing: \url{https://openrouter.ai/docs/guides/features/plugins/response-healing}}

Because execution traces can be large, we require summarisation models to support context windows of at least 1M tokens. We use models with input prices below \$1 per million tokens in the summarisation step, to keep the evaluation pipeline economically practical. 

We evaluate three summarisation and four judge models by measuring their agreement with human raters. For each model pair, we compute the mean pairwise Cohen's kappa between the automatic labels and the human annotations. \autoref{tab:llm_judge} reports the results.

\begin{table}[b]
    \centering
    \small
    \begin{tabular}{lrrr}
        \toprule
        & \multicolumn{3}{c}{\textbf{Summary model}} \\
        \cmidrule(lr){2-4}
        \textbf{Judge model} & Grok 4.1 Fast & Gemini 3 Flash & GPT 4.1 Nano \\
        \midrule
        Grok 4.1 Fast & -0.33 & 0.71 & 0.57 \\
        Gemini 3 Flash & -0.31 & 0.62 & 0.66 \\
        GPT 4.1 Nano & -0.11 & 0.49 & 0.19 \\
        Claude Sonnet 4.6 & -0.26 & \textbf{0.72} & 0.38 \\
        \bottomrule
    \end{tabular}
    \caption{Mean pairwise Cohen's kappa ($\kappa$) between human raters and the automatic evaluation pipeline. Columns denote the summary model and rows denote the judge model.}
    \label{tab:llm_judge}
\end{table}

Alignment varies substantially across model combinations. Along the main diagonal, where the same model performs both summarisation and judging, \texttt{Gemini 3 Flash} yields the strongest performance with $\kappa = 0.6241$. The best overall result is achieved when \texttt{Gemini 3 Flash} performs summarisation and \texttt{Claude Sonnet 4.6} performs judging, reaching $\kappa = 0.7234$. In contrast, every configuration that uses \texttt{grok-4.1-fast} for summarisation produces negative kappa values, regardless of the judge model. This pattern suggests that summarisation quality is a major determinant of downstream labelling reliability: weak summaries can erase or distort technical evidence before the judge sees it, while strong summaries improve the performance of multiple judge models.

These results show that modern LLMs can approximate human checkpoint labelling closely enough to support scalable benchmark evaluation, provided that the summarisation stage is chosen carefully. In operational terms, this pipeline reduces per-trace evaluation from multiple human hours to a few minutes, making repeated benchmarking of autonomous CTF agents feasible.

\section{Results} 
\label{sec:results}

\begin{table}[tb]
    \centering
    \begin{tabular}{l|cc|c}
Model & Tokens (M) & Steps & Completion \\
  \hline
  GPT-5.1 Codex Max & 31.9 & 57.0 & 35\% [29\%-44\%] \\
  MiniMax-M1 & 43.7 & 59.9 & 22\% [19\%-25\%] \\
  MiMo-V2 Flash & 58.1 & 59.7 & 20\% [19\%-22\%] \\
  Devstral 2512 & 56.0 & 53.2 & 21\% [16\%-27\%] \\
  Palmyra X5 & 20.6 & 42.7 & 20\% [14\%-25\%] \\
  Gemini 2.0 Flash & 28.7 & 48.9 & 16\% [13\%-18\%] \\
  Nova-2 Lite v1 & 46.4 & 55.4 & 13\% [12\%-16\%] \\
  Grok Code Fast 1 & 8.8 & 23.1 & 12\% [10\%-16\%] \\
  Qwen3-Next 80B & 16.6 & 31.0 & 12\% [10\%-13\%] \\
  Nemotron-3 Nano & 18.8 & 43.6 & 5\% [3\%-10\%] \\
    \end{tabular}
    \caption{Aggregated performance of evaluated models across all challenges. Tokens represent the average number of input and output tokens consumed per challenge in millions (M), and steps denote the average number of agent actions executed per run.}
    \label{tab:llm_res}
\end{table}

We evaluate ten LLM-based agents on the ten selected CTF challenges using the benchmark harness. Due to cost constraints, we run each model-challenge configuration three times and report the resulting average checkpoint completion. In pilot experiments, the run-to-run variance was generally low, making three repetitions a practical compromise between statistical rigour and evaluation cost. Each LLM runs with its default parameter configuration as defined by the model provider.

Table~\ref{tab:llm_res} summarises aggregate model performance across all challenges. Overall performance remains modest: the best-performing model, GPT-5.1 Codex Max, achieves 35\% average checkpoint completion, while only a small number of models exceed 20\%. A second tier of models—MiniMax-M1, Devstral 2512, MiMo-V2 Flash, and Palmyra X5—clusters between 20\% and 22\%, whereas the weakest model, Nemotron-3 Nano 30B, reaches only 5\%. Even the strongest systems therefore complete only about one third of the benchmark on average.

Token usage varies widely across models, ranging from $8.8$M tokens per run for \texttt{Grok Code Fast 1} to $58.1$M for \texttt{MiMo-V2 Flash}. However, higher token consumption does not reliably predict better outcomes. For example, \texttt{GPT-5.1 Codex Max} achieves the best completion score while using fewer tokens than several weaker models, including \texttt{MiniMax-M1}, \texttt{MiMo-V2 Flash}, and \texttt{Devstral 2512}. Conversely, \texttt{Grok Code Fast 1} is comparatively cheap in both tokens and steps, but this efficiency comes with relatively weak task performance. Taken together, these observations suggest that task success depends less on raw interaction volume than on the quality of reasoning and action selection within the available budget.

Average step counts cluster relatively tightly, typically between 40 and 60 actions per run, likely reflecting the imposed interaction limit. Consequently, variation in performance appears to arise less from differences in trajectory length than from the effectiveness of the actions taken within a similar number of opportunities.

\begin{table}[tb]
    \centering
    \begin{tabular}{l|cc|c}
  Challenge & Tokens (M) & Steps & Completion \\
  \hline
  Whitedoor & 27.2 & 45.3 & 24.6\% \\
  Quick & 32.0 & 46.4 & 23.5\% \\
  Sysadmin & 29.1 & 42.9 & 22.8\% \\
  Jan & 32.8 & 47.1 & 21.4\% \\
  Aria & 32.8 & 47.5 & 17.3\% \\
  Todd & 38.7 & 50.2 & 15.6\% \\
  Nexus & 35.3 & 51.2 & 15.2\% \\
  Helpdesk & 36.3 & 47.3 & 14.8\% \\
  Thirteen & 24.2 & 46.9 & 12.9\% \\
  Fuzzz & 41.2 & 50.1 & 10.1\% \\
    \end{tabular}
    \caption{Aggregated performance per challenge.}
    \label{tab:chal_res}
\end{table}

\autoref{tab:chal_res} shows the aggregated difficulty of individual challenges. Difficulty varies substantially across tasks. \texttt{Whitedoor}, \texttt{Quick}, and \texttt{SysAdmin} are the most solvable challenges in the suite. At the other end of the spectrum, \texttt{Fuzzz} is the hardest challenge, followed by \texttt{Thirteen}.

Challenge-level token usage also varies considerably. \texttt{Fuzzz}, \texttt{Todd}, \texttt{HelpDesk}, and \texttt{Nexus} require the largest average token budgets, whereas \texttt{Thirteen}, \texttt{Whitedoor}, and \texttt{SysAdmin} are comparatively lightweight. By contrast, step counts remain fairly stable across challenges, typically between $43$ and $51$ actions.

\section{Discussion}
Our findings contribute to the broader question of how effectively LLM-based agents can autonomously solve CTF challenges, and, how such progress can be evaluated in a rigorous and scalable manner. While the absolute task performance observed in \autoref{sec:results} remains limited, the results demonstrate that meaningful intermediate progress can nevertheless be detected automatically from agent interaction logs. This is an important prerequisite for studying autonomous offensive security systems in realistic environments, where binary success-or-failure evaluation is often too coarse to capture substantive differences in capability.

Challenges are hardest when they require non-standard discovery and multi-step state tracking rather than a familiar web foothold followed by privilege escalation. In particular, \texttt{Fuzzz}, \texttt{Thirteen}, and \texttt{Todd} are among the most difficult tasks, and all involve less conventional elements such as directory brute-forcing, SSH key handling, obfuscated disclosure, or service-specific exploitation. By contrast, challenges such as \texttt{Whitedoor}, \texttt{Quick}, and \texttt{SysAdmin} follow a more standard web-to-privilege-escalation pattern and are completed more often. This suggests that current agents perform better on common exploitation workflows, but struggle when they must combine broad enumeration with unusual artefacts or less common attack paths.

A central contribution of this work is the use of log-based labelling to infer challenge progress from recorded trajectories. In particular, the method appears capable of identifying semantically meaningful stages of problem-solving, such as reconnaissance, artefact discovery, and partial exploitation, without requiring manual inspection of every agent run.

At the same time, the quality of the resulting labels depends directly on the quality of the manually defined milestone set. Since these milestones are derived from writeups, any incompleteness, ambiguity, or conciseness in the writeup can propagate into the evaluation. A writeup may omit intermediate reasoning steps, collapse multiple actions into a single narrative transition, or describe only one of several viable solution paths, causing the extracted milestones to underrepresent legitimate alternative forms of progress. When a challenge has multiple solution paths, the checkpoint rubric should reflect this by defining separate checkpoint sequences for each path. We did not observe such cases in the manual evaluation, and the writeups used here appeared sufficiently detailed to support consistent milestone extraction.

The results also illustrate why partial scoring is especially important in this domain. Many agents make measurable early progress, particularly on reconnaissance and enumeration steps, yet fail to complete later stages that require longer-horizon planning, adaptation to unexpected outputs, or more creative exploitation. 
This is particularly valuable for analysing emerging agentic capabilities, where full task completion may still be rare but intermediate progress is already informative.

An additional practical consideration concerns the cost of human annotation. Even though the proposed pipeline reduces the need for manual scoring of every individual run, constructing or validating challenge labels still requires substantial human effort. Raters must understand both the challenge itself and the associated writeup in sufficient depth to identify meaningful milestones and assess whether inferred progress is plausible. This creates a high initial startup cost and requires domain expertise in security and exploitation, which makes the task inaccessible to lay annotators. As a result, although the framework improves scalability relative to fully manual evaluation, it does not eliminate expert labour altogether. Future work should therefore consider how milestone extraction, writeup alignment, and challenge decomposition might be further standardised or partially automated.

We observed an infrastructure failure that reinforces the importance of strong isolation. In one live run, shutting down the target VM did not terminate the agent, which continued operating on the host, scanned the local network, and began probing other active systems and services. We stopped the process after detecting the unauthorised activity and patched the issue.

\subsection{Agent Improvement Opportunities}
\label{subsec:agent_improvement}
The failure patterns observed during manual evaluation indicate that the principal bottlenecks for current agents are not related to basic tool use or knowledge of the utilities in Kali Linux. Most models can issue commands, inspect files, and gather surface-level information using the available toolchain. However, many appear to struggle when success depends on synthesising observations over multiple steps, revising an unsuccessful plan, or identifying less obvious exploitation opportunities.

We note that agents tend to lack perseverance. While some agents give up prematurely, we find that most progress is made in the initial ~20 steps, after which the agents tend to get lost. Some gain initial user access to the challenge VM, and then look for the flag without escalating their privileges to root. Others continuously repeat the same failing commands or start over multiple times. This type of long-horizon planning still seems to be challenging for all tested agents.

Agents tend to struggle with the fact that the Kali environment persists throughout a challenge episode rather than being reset between steps. So they rarely store intermediate results and tend to redo many steps.

\subsection{Threats to validity}
\label{subsec:threats}

\textit{Internal validity.}
A potential threat is the use of SmolAgents as a common framework for all models. Some models may perform better under different prompting schemes or agent architectures. We accept this limitation, as a shared framework is necessary for comparability and implementing model-specific systems for each of the selected models is not practical. A second threat is data leakage: some challenges or writeups may have appeared in model training data, potentially inflating performance. This cannot be ruled out, particularly for closed-data models with undisclosed training data and cut-off dates. 
The web search tool is a possible source of data leakage, but throughout the manual evaluation we found that the filters removed any information directly related to the challenges.

Finally, while the non-determinism of LLMs ideally requires a large number of trials to achieve statistical significance, we limited our evaluation to three executions per model for each challenge due to time and financial constraints. Although aggregating results across these three passes helps mitigate variance, this remains a limited sample size and may not fully capture the complete range of model behaviours or edge-case failures.

\textit{External validity.}
DeepRed includes only ten models and ten challenges, which limits generalisability. We mitigate this by selecting a diverse set of models and tasks within a constrained evaluation budget, and by designing the benchmark to be extensible. The challenges are also relatively easy, which improves tractability for current agents but means the results may not transfer to harder or more realistic offensive security settings.

The system prompt is a known confound in agent benchmarking~\cite{yao2022react}, as different prompt formulations can produce meaningfully different absolute scores. We mitigate this partially by applying the same prompt across all models, which preserves the validity of relative comparisons between systems. 

\textit{Construct validity.}
CTF performance is only a proxy for broader cybersecurity capability. CTFs are useful because they are controlled, reproducible, and safe, but they do not capture many real-world factors such as stealth, persistence, lateral movement, or interaction with defenders.

\subsection{Future work}
As discussed in \autoref{subsec:threats}, our evaluation uses SmolAgents as a shared backbone to ensure comparability, but this may not reflect the ceiling performance of models better suited to other architectures. Promising alternatives include \textit{LangGraph} for finer tool calling control, \textit{AutoGen} for multi-agent collaboration, and \textit{OpenHands} for terminal-heavy workflows. Future work should examine whether model rankings are stable across frameworks or whether architecture-model pairings produce meaningfully different outcomes.

Many failures observed in this work, discussed in \autoref{subsec:agent_improvement}, stem not from a lack of domain knowledge, but from poor long-horizon planning, weak memory usage, and insufficient adaptation to failed attempts. Future work could explore hierarchical planning agents, explicit memory systems, and self-reflection loops as targeted remedies. Multi-agent architectures, where specialised agents handle different stages independently, represent a promising direction for decomposing the long task horizons that current single-agent systems struggle to sustain.

A further direction for future work is to improve the robustness and generality of the automatic labelling pipeline. Checkpoint labels currently depend on manually derived rubrics extracted from public writeups. Future work could investigate methods for partially automating checkpoint extraction from writeups. A similar evaluation pipeline could be applied to other long-horizon tasks with public guides.

\subsection{Reproducibility}
We publicly release \textbf{DeepRed} together with challenge configuration files, evaluation scripts, scoring definitions, evaluation logs, and summaries.\footnote{Replication Package: \url{https://doi.org/10.5281/zenodo.19386535}}\footnote{Github:  \url{https://github.com/AISE-TUDelft/DeepRed-LLMAgent}}

We design the dataset to remain extensible: researchers and practitioners can add new challenges by supplying a VM image and a corresponding checkpoint rubric. More models can be evaluated on the provided core set using any OpenAI-compatible API. We have consumed around \$450 of OpenRouter credits throughout development. A benchmark sweep over the selected models and challenges costs around \$70 per run set as of April 2026.

\section{Conclusion}
In this work, we introduced \textbf{DeepRed}, a reproducible benchmark framework for evaluating LLM-based agents in realistic CTF environments. Across ten challenges and ten models, our results show that current agents can make partial progress on challenging tasks but remain far from robust end-to-end autonomy. Capturing partial progress is important because binary solve rates can hide substantial differences in capability on multi-step tasks. DeepRed enables reproducible capability tracking and future research on agentic failure analysis and evaluation design, and we encourage the community to build on and extend this benchmark as agentic systems continue to evolve.

\bibliographystyle{ACM-Reference-Format}
\bibliography{bibliography}

@article{hou2023large,
  title={Large language models for software engineering: A systematic literature review},
  author={Hou, Xinyi and Zhao, Yanjie and Liu, Yue and Yang, Zhou and Wang, Kailong and Li, Li and Luo, Xiapu and Lo, David and Grundy, John and Wang, Haoyu},
  journal={arXiv preprint arXiv:2308.10620},
  year={2023}
}

@inproceedings{izadi2024language,
  title={Language models for code completion: A practical evaluation},
  author={Izadi, Maliheh and Katzy, Jonathan and Van Dam, Tim and Otten, Marc and Popescu, Razvan Mihai and Van Deursen, Arie},
  booktitle={Proceedings of the IEEE/ACM 46th International Conference on Software Engineering},
  pages={1--13},
  year={2024}
}

@misc{siddiq2024generate,
      title={Generate and Pray: Using SALLMS to Evaluate the Security of LLM Generated Code}, 
      author={Mohammed Latif Siddiq and Joanna C. S. Santos and Sajith Devareddy and Anna Muller},
      year={2024},
      eprint={2311.00889},
      archivePrefix={arXiv},
      primaryClass={cs.SE},
      url={https://arxiv.org/abs/2311.00889}, 
}

@article{pearce2025asleep,
  title={Asleep at the keyboard? assessing the security of github copilot’s code contributions},
  author={Pearce, Hammond and Ahmad, Baleegh and Tan, Benjamin and Dolan-Gavitt, Brendan and Karri, Ramesh},
  journal={Communications of the ACM},
  volume={68},
  number={2},
  pages={96--105},
  year={2025},
  publisher={ACM New York, NY, USA}
}

@article{hasanov2024application,
  title={Application of large language models in cybersecurity: A systematic literature review},
  author={Hasanov, Ismayil and Virtanen, Seppo and Hakkala, Antti and Isoaho, Jouni},
  journal={IEEE access},
  volume={12},
  pages={176751--176778},
  year={2024},
  publisher={IEEE}
}

@article{brundage2018malicious,
  title={The malicious use of artificial intelligence: Forecasting, prevention, and mitigation},
  author={Brundage, Miles and Avin, Shahar and Clark, Jack and Toner, Helen and Eckersley, Peter and Garfinkel, Ben and Dafoe, Allan and Scharre, Paul and Zeitzoff, Thomas and Filar, Bobby and others},
  journal={arXiv preprint arXiv:1802.07228},
  year={2018}
}

@article{alkaswan2025code,
  title={Code red! on the harmfulness of applying off-the-shelf large language models to programming tasks},
  author={Al-Kaswan, Ali and Deatc, Sebastian and Ko{\c{c}}, Beg{\"u}m and van Deursen, Arie and Izadi, Maliheh},
  journal={Proceedings of the ACM on Software Engineering},
  volume={2},
  number={FSE},
  pages={2477--2499},
  year={2025},
  publisher={ACM New York, NY, USA}
}

@article{shao2024nyu,
  title={Nyu ctf bench: A scalable open-source benchmark dataset for evaluating llms in offensive security},
  author={Shao, Minghao and Jancheska, Sofija and Udeshi, Meet and Dolan-Gavitt, Brendan and Milner, Kimberly and Chen, Boyuan and Yin, Max and Garg, Siddharth and Krishnamurthy, Prashanth and Khorrami, Farshad and others},
  journal={Advances in Neural Information Processing Systems},
  volume={37},
  pages={57472--57498},
  year={2024}
}

@inproceedings{thaqi2024leveraging,
  title={Leveraging ai for ctf challenge optimization},
  author={Thaqi, Alba and Musa, Arbena and Rexha, Blerim},
  booktitle={2024 5th International Conference on Communications, Information, Electronic and Energy Systems (CIEES)},
  pages={1--5},
  year={2024},
  organization={IEEE}
}

@inproceedings{deng2024pentestgpt,
  title={$\{$PentestGPT$\}$: Evaluating and harnessing large language models for automated penetration testing},
  author={Deng, Gelei and Liu, Yi and Mayoral-Vilches, V{\'\i}ctor and Liu, Peng and Li, Yuekang and Xu, Yuan and Zhang, Tianwei and Liu, Yang and Pinzger, Martin and Rass, Stefan},
  booktitle={33rd USENIX Security Symposium (USENIX Security 24)},
  pages={847--864},
  year={2024}
}

@inproceedings{
  abramovich2025enigma,
  title={En{IGMA}: Interactive Tools Substantially Assist {LM} Agents in Finding Security Vulnerabilities},
  author={Talor Abramovich and Meet Udeshi and Minghao Shao and Kilian Lieret and Haoran Xi and Kimberly Milner and Sofija Jancheska and John Yang and Carlos E Jimenez and Farshad Khorrami and Prashanth Krishnamurthy and Brendan Dolan-Gavitt and Muhammad Shafique and Karthik R Narasimhan and Ramesh Karri and Ofir Press},
  booktitle={Forty-second International Conference on Machine Learning},
  year={2025},
  url={https://openreview.net/forum?id=Of3wZhVv1R}
}

@inproceedings{zhu2025cvebench,
  title={CVE-Bench: A Benchmark for AI Agents’ Ability to Exploit Real-World Web Application Vulnerabilities},
  author={Zhu, Yuxuan and Kellermann, Antony and Bowman, Dylan and Li, Philip and Gupta, Akul and Danda, Adarsh and Fang, Richard and Jensen, Conner and Ihli, Eric and Benn, Jason and others},
  booktitle={International Conference on Machine Learning},
  pages={79850--79867},
  year={2025},
  organization={PMLR}
}

@article{conceiccao2025evaluation,
  title={Evaluation of the maturity of LLMs in the cybersecurity domain: T. Concei{\c{c}}{\~a}o, N. Cruz},
  author={Concei{\c{c}}{\~a}o, Tiago and Cruz, Nuno},
  journal={International Journal of Information Security},
  volume={24},
  number={5},
  pages={197},
  year={2025},
  publisher={Springer}
}

@article{combe2016docker,
  title={To Docker or not to Docker: A security perspective},
  author={Combe, Theo and Martin, Antony and Di Pietro, Roberto},
  journal={IEEE Cloud Computing},
  volume={3},
  number={5},
  pages={54--62},
  year={2016},
  publisher={IEEE}
}

@misc{wang2024executable,
  title={Executable Code Actions Elicit Better LLM Agents}, 
  author={Xingyao Wang and Yangyi Chen and Lifan Yuan and Yizhe Zhang and Yunzhu Li and Hao Peng and Heng Ji},
  year={2024},
  eprint={2402.01030},
  archivePrefix={arXiv},
  primaryClass={cs.CL}
}

@inproceedings{yao2022react,
  title={React: Synergizing reasoning and acting in language models},
  author={Yao, Shunyu and Zhao, Jeffrey and Yu, Dian and Du, Nan and Shafran, Izhak and Narasimhan, Karthik R and Cao, Yuan},
  booktitle={The eleventh international conference on learning representations},
  year={2022}
}

@article{schick2023toolformer,
  title={Toolformer: Language models can teach themselves to use tools},
  author={Schick, Timo and Dwivedi-Yu, Jane and Dess{\`\i}, Roberto and Raileanu, Roberta and Lomeli, Maria and Hambro, Eric and Zettlemoyer, Luke and Cancedda, Nicola and Scialom, Thomas},
  journal={Advances in neural information processing systems},
  volume={36},
  pages={68539--68551},
  year={2023}
}

@inproceedings{park2023generative,
  title={Generative agents: Interactive simulacra of human behavior},
  author={Park, Joon Sung and O'Brien, Joseph and Cai, Carrie Jun and Morris, Meredith Ringel and Liang, Percy and Bernstein, Michael S},
  booktitle={Proceedings of the 36th annual acm symposium on user interface software and technology},
  pages={1--22},
  year={2023}
}

@article{jimenez2023swe,
  title={Swe-bench: Can language models resolve real-world github issues?},
  author={Jimenez, Carlos E and Yang, John and Wettig, Alexander and Yao, Shunyu and Pei, Kexin and Press, Ofir and Narasimhan, Karthik},
  journal={arXiv preprint arXiv:2310.06770},
  year={2023}
}

@inproceedings{ba2024covernexus,
  title={CoverNexus: Multi-agent LLM System for Automated Code Coverage Enhancement},
  author={Ba, Thiem Nguyen and Thanh, Binh Nguyen and Tran, Viet-Trung},
  booktitle={International Symposium on Information and Communication Technology},
  pages={472--484},
  year={2024},
  organization={Springer}
}

@article{yang2024swe,
  title={Swe-agent: Agent-computer interfaces enable automated software engineering},
  author={Yang, John and Jimenez, Carlos E and Wettig, Alexander and Lieret, Kilian and Yao, Shunyu and Narasimhan, Karthik and Press, Ofir},
  journal={Advances in Neural Information Processing Systems},
  volume={37},
  pages={50528--50652},
  year={2024}
}

@inproceedings{ahmad2025using,
  title={Using reinforcement learning for security testing: A systematic mapping study},
  author={Ahmad, Tanwir and Butkovic, Matko and Truscan, Dragos},
  booktitle={2025 IEEE International Conference on Software Testing, Verification and Validation Workshops (ICSTW)},
  pages={208--216},
  year={2025},
  organization={IEEE}
}

@Misc{smolagents,
  title =        {`smolagents`: a smol library to build great agentic systems.},
  author =       {Aymeric Roucher and Albert Villanova del Moral and Thomas Wolf and Leandro von Werra and Erik Kaunismäki},
  howpublished = {\url{https://github.com/huggingface/smolagents}},
  year =         {2025}
}

@inproceedings{zhangcybench,
  title={Cybench: A Framework for Evaluating Cybersecurity Capabilities and Risks of Language Models},
  author={Zhang, Andy K and Perry, Neil and Dulepet, Riya and Ji, Joey and Menders, Celeste and Lin, Justin W and Jones, Eliot and Hussein, Gashon and Liu, Samantha and Jasper, Donovan Julian and others},
  booktitle={The Thirteenth International Conference on Learning Representations},
  year={2025}
}
\end{document}